\PassOptionsToPackage{table}{xcolor}
\documentclass[10pt, a4paper]{article}
\usepackage[table,xcdraw]{xcolor}
\usepackage{lrec-coling2024} 
\usepackage{multirow}
\usepackage{graphicx}
\NewDocumentCommand\emo{}{\includegraphics[scale=0.1]{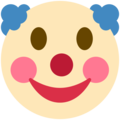}}
\usepackage{algorithm}
\usepackage{algorithmic}

\title{Examining Temporalities on Stance Detection Towards COVID-19 Vaccination}

\name{Yida Mu,
    Mali Jin,
    Kalina Bontcheva,
    Xingyi Song} 

\address{Department of Computer Science, The University of Sheffield\\
         \{y.mu, m.jin, k.bontcheva, x.song\}@sheffield.ac.uk\\}

\abstract{
Previous studies have highlighted the importance of vaccination as an effective strategy to control the transmission of the COVID-19 virus. It is crucial for policymakers to have a comprehensive understanding of the public's stance towards vaccination on a large scale. However, attitudes towards COVID-19 vaccination, such as pro-vaccine or vaccine hesitancy, have evolved over time on social media. Thus, it is necessary to account for possible temporal shifts when analysing these stances. This study aims to examine the impact of temporal concept drift on stance detection towards COVID-19 vaccination on Twitter. To this end, we evaluate a range of transformer-based models using chronological (splitting the training, validation, and test sets in order of time) and random splits (randomly splitting these three sets) of social media data. \textcolor{black}{Our findings reveal significant discrepancies in model performance between random and chronological splits in several existing COVID-19-related datasets; specifically, chronological splits significantly reduce the accuracy of stance classification.} Therefore, real-world stance detection approaches need to be further refined to incorporate temporal factors as a key consideration. 
 \\ \newline  }

\begin{document}

\maketitleabstract

\section{Introduction}
The COVID-19 pandemic has had a profound impact on global health and has resulted in considerable social and economic disruption \citep{ciotti2020covid}. 
Promoting vaccination has been statistically recognised as a vital tactic in curbing the spread of the COVID-19 virus and reducing the burden on healthcare systems \citep{lopez2021effectiveness}. However, attitudes towards COVID-19 vaccination have varied, with some individuals expressing hesitation and resistance \citep{cotfas2021longest,poddar2022winds}. 
These concerns are caused by factors including side effects, conspiracy theories, and distrust of healthcare authorities \citep{CAVES}. To promote vaccine uptake, it is important to understand the factors that contribute to hesitant or negative attitudes towards COVID-19 vaccination on a large scale \citep{mu2023vaxxhesitancy}. 
Recently, there has been a growing interest in using supervised machine learning approaches to automatically detect users' stance towards COVID-19 vaccination on social media \citep{di2022vaccineu,glandt2021stance,chen2022multilingual}. \footnote{Accepted at LREC-COLING 2024.}

Temporal concept drift refers to the phenomenon in NLP where the statistical properties of a dataset change over time such as the distribution of topics \citep{huang2019neural}.  \textcolor{black}{It is particularly relevant in applications where data is collected over extended periods, such as in financial forecasting or tasks based on social media data \citep{xing2018natural,alkhalifa2023building,hu-etal-2023-learn}.} It can also lead to the degradation of model performance due to the temporal variation of textual content in a static dataset \citep{mu2023s,mu2023examining2}. \textcolor{black}{The impact of temporal concept drift has been investigated in several domains including topic classification \cite{chalkidis2022improved}, rumour detection \cite{mu2023s}, gender equality \citep{alkhalifa2021opinions}, and hate speech detection \cite{florio2020time,jin2023examining}.} \textcolor{black}{Moreover, \cite{alkhalifa2023building} evaluated how different factors of datasets and models affect the performance over time across different classification tasks.} 

However, temporal aspects have not been studied in stance detection concerning COVID-19 vaccination. Furthermore, the underlying distribution of users' stances may change over time due to factors such as evolving political agendas and emerging viral variants, making it necessary to take the temporal factor into account.

\begin{table*}[!t]
\centering
\tiny
\resizebox{\textwidth}{!}{%
\begin{tabular}{|c|c|c|c|c|}
\hline
\textbf{Dataset} &
  \multicolumn{1}{c|}{\textbf{Time}} & 
  \textbf{Tweets} &
  \multicolumn{1}{c|}{\textbf{Labels}} &
  \textbf{Language} \\ \hline
\citet{cotfas2021longest} & Nov 2020 $\sim$Dec 2020         & 2,792   & in favour, against, neutral      & en                 \\ \hline
\citet{poddar2022winds} & Jan  2020 $\sim$March 2021         & 1,700  & in favour, against, neutral       & en                 \\ \hline
\citet{mu2023vaxxhesitancy}   & Nov 2020 $\sim$April 2022   & 3,101  & pro, anti, hesitancy, irrelevant   & en                 \\ \hline
\citet{chen2022multilingual}   & Jan  2020 $\sim$March 2021  & 17,934  & pos, neg, neutral, off-topic       & fr, de, en \\ \hline
\citet{di2022vaccineu}   & Nov 2020 $\sim$ June 2021        & 3,101  & in favour, against, neutral   & es, de, it                 \\ \hline
\end{tabular}%
}
\caption{Dataset statistics.}
\label{tab:related_work}
\end{table*}

This study examines the temporal concept drift in stance detection towards COVID-19 vaccination on Twitter for the first time.\footnote{Our code: \url{https://github.com/YIDAMU/COVID_Temporalites}} Specifically, we focus on the following research questions:
\begin{itemize}

    \item \textbf{\textit{Q1:}} Does temporal concept drift exert a significant affect on COVID-19 vaccine stance detection, as previously seen in other domains (e.g., rumour, legal, biomedical, etc.)?
    \item \textbf{\textit{Q2:}} How does the model's performance differ across multiple languages?
    \item \textbf{\textit{Q3:}}  Can domain adaptation approaches be employed to mitigate the temporal concept drift impact on stance classification?
    \item \textbf{\textit{Q4:}} Does the semantic variation between the training and test sets lead to a degradation or improvement of the model's predictive performance? 
\end{itemize}

To achieve this, we (i) evaluate five publicly available monolingual and multilingual datasets, (ii) conduct a set of controlled experiments by evaluating various transformer-based pretrained language models (PLMs) using \textit{chronological} and \textit{random splits} and (iii) perform correlation tests to examine the relationship (i.e., positive or negative) between the model predictive performance and the disparity between the two subsets (i.e., training and test sets).

\section{Experimental Setup}
\subsection{Datasets}
\textcolor{black}{We use three datasets}
\footnote{(i) \url{https://github.com/liviucotfas/covid-19-vaccination-stance-detection}; (ii) \url{https://github.com/sohampoddar26/covid-vax-stance}; (iii) \url{https://zenodo.org/records/7601328}} in English \citep{cotfas2021longest,poddar2022winds,mu2023vaxxhesitancy} and two datasets\footnote{(iv) \url{https://zenodo.org/records/5851407}; (v) \url{https://github.com/datasciencepolimi/vaccineu}} in multiple languages \citep{chen2022multilingual,di2022vaccineu}. 
These datasets 
adhere to the FAIR principles (i.e., Findable, Accessible, Interoperable and Re-usable) \citep{wilkinson2016fair}. 
More details (e.g. sources, annotation details, label definitions) can be found in the original articles.
Differences in specifications between these datasets are shown in Table \ref{tab:related_work}.

\subsection{Data Splits}
We aim to conduct a set of controlled experiments to explore the impact of temporality on the accuracy of stance classifiers regarding COVID-19 vaccination. To this end, we evaluate two data split strategies:
\begin{itemize}
    \item \textbf{Chronological Splits}
Following \citet{mu2023s}, all datasets are sorted chronologically and subsequently divided into a training set (70\% earliest data), a validation set (10\% data after training set and before test set) and a test set (20\% latest data). Note that the three subsets do not overlap temporally.
    \item \textbf{Random Splits}
\textcolor{black}{We randomly split all datasets using a stratified 5-fold cross-validation method, ensuring that the class proportions in each fold mirror those in the original dataset. Additionally, the ratio of training to validation to testing matches that of the chronological splits.}
\end{itemize}




\begin{table*}[!t]
\tiny
\resizebox{\textwidth}{!}{%
\begin{tabular}{|c|l|llll|llll|llll|}
\hline
 &
  \multicolumn{1}{c|}{} &
  \multicolumn{4}{c|}{\textbf{\citet{cotfas2021longest}}} &
  \multicolumn{4}{c|}{\textbf{\citet{poddar2022winds}}} &
  \multicolumn{4}{c|}{\textbf{\citet{mu2023vaxxhesitancy}}} \\ \cline{3-14} 
\multirow{-2}{*}{\textbf{Model}} &
  \multicolumn{1}{c|}{\multirow{-2}{*}{\textbf{Splits}}} &
  \multicolumn{1}{c|}{P} &
  \multicolumn{1}{c|}{R} &
  \multicolumn{1}{c|}{F1} &
  Acc &
  \multicolumn{1}{c|}{P} &
  \multicolumn{1}{c|}{R} &
  \multicolumn{1}{c|}{F1} &
  Acc &
  \multicolumn{1}{c|}{P} &
  \multicolumn{1}{c|}{R} &
  \multicolumn{1}{c|}{F1} &
  Acc \\ \hline
 &
  \textit{Random} &
  \multicolumn{1}{l|}{79.2} &
  \multicolumn{1}{l|}{79.3} &
  \multicolumn{1}{l|}{79.2} &
  79.4 &
  \multicolumn{1}{l|}{50.8} &
  \multicolumn{1}{l|}{47.6} &
  \multicolumn{1}{l|}{43.5} &
  58.4 &
  \multicolumn{1}{l|}{54.6} &
  \multicolumn{1}{l|}{54.6} &
  \multicolumn{1}{l|}{54.4} &
  54.5 \\ \cline{2-14} 
\multirow{-2}{*}{\textbf{BERT}} &
  \textit{Chronological} &
  \multicolumn{1}{l|}{\cellcolor[HTML]{C0C0C0}62.9} &
  \multicolumn{1}{l|}{\cellcolor[HTML]{C0C0C0}64.6} &
  \multicolumn{1}{l|}{\cellcolor[HTML]{C0C0C0}63.3} &
  \cellcolor[HTML]{C0C0C0}74.5 &
  \multicolumn{1}{l|}{46.9} &
  \multicolumn{1}{l|}{45.4} &
  \multicolumn{1}{l|}{41.1} &
  57.4 &
  \multicolumn{1}{l|}{52.9} &
  \multicolumn{1}{l|}{53.2} &
  \multicolumn{1}{l|}{52.5} &
  53.0 \\ \hline
 &
  \textit{Random} &
  \multicolumn{1}{l|}{85.8} &
  \multicolumn{1}{l|}{85.1} &
  \multicolumn{1}{l|}{85.3} &
  85 &
  \multicolumn{1}{l|}{71.2} &
  \multicolumn{1}{l|}{68.6} &
  \multicolumn{1}{l|}{69.3} &
  71.0 &
  \multicolumn{1}{l|}{67.3} &
  \multicolumn{1}{l|}{67.2} &
  \multicolumn{1}{l|}{67.2} &
  67.2 \\ \cline{2-14} 
\multirow{-2}{*}{\textbf{\begin{tabular}[c]{@{}c@{}}COVID\\ BERT\end{tabular}}} &
  \textit{Chronological} &
  \multicolumn{1}{l|}{\cellcolor[HTML]{C0C0C0}74.2} &
  \multicolumn{1}{l|}{\cellcolor[HTML]{C0C0C0}75} &
  \multicolumn{1}{l|}{\cellcolor[HTML]{C0C0C0}\textbf{74.5}} &
  \cellcolor[HTML]{C0C0C0}80.4 &
  \multicolumn{1}{l|}{67.7} &
  \multicolumn{1}{l|}{68.0} &
  \multicolumn{1}{l|}{\textbf{67.6}} &
  70.5 &
  \multicolumn{1}{l|}{67.6} &
  \multicolumn{1}{l|}{68.0} &
  \multicolumn{1}{l|}{\textbf{67.7}} &
  68.3 \\ \hline
\multicolumn{1}{|l|}{} &
  \textit{Random} &
  \multicolumn{1}{l|}{85.4} &
  \multicolumn{1}{l|}{85.3} &
  \multicolumn{1}{l|}{85.4} &
  85.1 &
  \multicolumn{1}{l|}{71.2} &
  \multicolumn{1}{l|}{66.4} &
  \multicolumn{1}{l|}{67.5} &
  69.9 &
  \multicolumn{1}{l|}{67.7} &
  \multicolumn{1}{l|}{67.7} &
  \multicolumn{1}{l|}{67.6} &
  67.6 \\ \cline{2-14} 
\multicolumn{1}{|l|}{\multirow{-2}{*}{\textbf{\begin{tabular}[c]{@{}l@{}}VAXX\\ BERT\end{tabular}}}} &
  \textit{Chronological} &
  \multicolumn{1}{l|}{\cellcolor[HTML]{C0C0C0}71.8} &
  \multicolumn{1}{l|}{\cellcolor[HTML]{C0C0C0}71.2} &
  \multicolumn{1}{l|}{\cellcolor[HTML]{C0C0C0}70.7} &
  \cellcolor[HTML]{C0C0C0}79.3 &
  \multicolumn{1}{l|}{65.5} &
  \multicolumn{1}{l|}{63.7} &
  \multicolumn{1}{l|}{64.1} &
  66.8 &
  \multicolumn{1}{l|}{67.5} &
  \multicolumn{1}{l|}{67.3} &
  \multicolumn{1}{l|}{67.4} &
  68.2 \\ \hline
\end{tabular}%
}
\caption{Model predictive performance on mono-lingual datasets. Cells in \textit{Grey} denotes that the classifier trained on random splits performs significantly better than chronological splits ($p < 0.05$, $t$-test). The smallest performance drop (or increase) using chronological splits is in bold.}
\label{tab:results_monolingual}
\end{table*}

\begin{table*}[!t]
\centering
\small
\footnotesize
\begin{tabular}{|c|l|llll|llll|}
\hline
 &
  \multicolumn{1}{c|}{} &
  \multicolumn{4}{c|}{\textbf{\citet{di2022vaccineu}}} &
  \multicolumn{4}{c|}{\textbf{\citet{chen2022multilingual}}} \\ \cline{3-10} 
\multirow{-2}{*}{\textbf{Model}} &
  \multicolumn{1}{c|}{\multirow{-2}{*}{\textbf{Splits}}} &
  \multicolumn{1}{c|}{P} &
  \multicolumn{1}{c|}{R} &
  \multicolumn{1}{c|}{F1} &
  Acc &
  \multicolumn{1}{c|}{P} &
  \multicolumn{1}{c|}{R} &
  \multicolumn{1}{c|}{F1} &
  Acc \\ \hline
 &
  \textit{Random Splits} &
  \multicolumn{1}{l|}{42.2} &
  \multicolumn{1}{l|}{41.6} &
  \multicolumn{1}{l|}{41.6} &
  52.8 &
  \multicolumn{1}{l|}{63.5} &
  \multicolumn{1}{l|}{62.1} &
  \multicolumn{1}{l|}{62.7} &
  75.4 \\ \cline{2-10} 
\multirow{-2}{*}{\textbf{\begin{tabular}[c]{@{}c@{}}XML-BERT\end{tabular}}} &
  \textit{Chronological Splits} &
  \multicolumn{1}{l|}{42.4} &
  \multicolumn{1}{l|}{41.8} &
  \multicolumn{1}{l|}{\textbf{42.0}} &
  52.9 &
  \multicolumn{1}{l|}{\cellcolor[HTML]{C0C0C0}60.6} &
  \multicolumn{1}{l|}{\cellcolor[HTML]{C0C0C0}60.6} &
  \multicolumn{1}{l|}{\cellcolor[HTML]{C0C0C0}57.9} &
  \cellcolor[HTML]{C0C0C0}73.4 \\ \hline
 &
  \textit{Random Splits} &
  \multicolumn{1}{l|}{45.8} &
  \multicolumn{1}{l|}{43.9} &
  \multicolumn{1}{l|}{44.2} &
  55.1 &
  \multicolumn{1}{l|}{65.0} &
  \multicolumn{1}{l|}{64.0} &
  \multicolumn{1}{l|}{64.5} &
  77.6 \\ \cline{2-10} 
\multirow{-2}{*}{\textbf{\begin{tabular}[c]{@{}c@{}}XML-RoBERTa\end{tabular}}} &
  \textit{Chronological Splits} &
  \multicolumn{1}{l|}{45.2} &
  \multicolumn{1}{l|}{43.4} &
  \multicolumn{1}{l|}{43.7} &
  55.0 &
  \multicolumn{1}{l|}{\cellcolor[HTML]{C0C0C0}62.4} &
  \multicolumn{1}{l|}{\cellcolor[HTML]{C0C0C0}61.9} &
  \multicolumn{1}{l|}{\cellcolor[HTML]{C0C0C0}\textbf{62}} &
  \cellcolor[HTML]{C0C0C0}75.5 \\ \hline
\end{tabular}%
\caption{Model predictive performance on multilingual datasets. Cells in \textit{Grey} denotes that the classifier trained on random splits performs significantly better than chronological splits ($p < 0.05$, $t$-test). The smallest performance drop (or increase) using chronological splits is in bold.} 
\vspace{-4mm}
\label{tab:results_multilingual}
\end{table*}

\subsection{Models}
We evaluate various transformer-based PLMs. Following \citet{devlin2019bert}, we fine-tune these PLMs by adding a fully-connected layer on top of the transformer architecture. We consider the special token `[CLS]' as the tweet-level representation.
\textbf{Mono-lingual PLMs} 
To represent tweets in English, we consider vanilla BERT and two domain-adapted PLMs: 

\begin{itemize}
    \item \textbf{BERT} \citep{devlin2019bert} is trained on a large corpus including Wikipedia articles and English Books-Corpus.
    \item \textbf{COVID-BERT} \citep{muller2020covid} is a specialised version of BERT model that
    has been further pre-trained on a large corpus of COVID-19 related texts to improve the performance of related downstream tasks \citep{cotfas2021longest,poddar2022winds}.\footnote{\small \url{https://huggingface.co/digitalepidemiologylab/covid-twitter-bert-v2}} 
    \item \textbf{Vaccine-BERT} \citep{mu2023vaxxhesitancy} is a domain adapted BERT model for automatically detecting COVID-19 vaccine stance.\footnote{\url{https://huggingface.co/GateNLP/covid-vaccine-twitter-bert}} 
    It has been further pre-trained on a large dataset of tweets related to \textit{COVID-19 vaccines} based on the COVID-BERT model to improve its ability to accurately classify the vaccine stance.
\end{itemize}

\textbf{Multilingual PLMs} 
To represent tweets in multilingual, we consider two strong cross-lingual PLMs:
\begin{itemize}
    \item \textbf{XLM-BERT} \citep{devlin2019bert} is a multilingual version of BERT that has been pre-trained on texts from over 100 multiple languages. 
    \item \textbf{XLM-RoBERTa} \citep{conneau2020unsupervised} 
    is trained to reconstruct a sentence in one language from a corrupted version of the sentence in another language, which has been shown highly effective for multilingual NLP tasks such as cross-lingual stance classification \citep{chen2022multilingual}.
\end{itemize}

\subsection{Training \& Evaluation}
For all datasets, we keep the original setup (i.e., multi-class classification task). We pre-process the tweets from all datasets by (i) lowercasing and (ii) replacing @user\_name and hyperlinks with special tokens i.e., `@USER' and `HTTPURL' respectively. 

All models are trained on the training set, while model tuning and selection are based on the validation loss observed at each training epoch. Subsequently, the predictive performance of the model is assessed on the test set.
We run all models five times with varying random seeds to ensure consistency and report the average Accuracy, Precision, Recall, and macro-F1 scores. 
\textcolor{black}{
For all PLMs, we set learning rate as 2e-5, batch size as 16, and max number of input tokens as 256. 
All experiments are performed on a NVIDIA Titan RTX GPU with 24 GB memory.}

\section{Analysis}
In this section, we present a detailed similarity analysis at different levels (e.g., token and topic) and conduct an error analysis based on the output of the model.
\subsection{Text Similarity}
Our aim is to investigate whether a decrease in model predictive performance occurs due to variations between the two subsets used for training and testing, and whether the difference in performance lessens as the datasets become more similar to each other.
Following \citet{kochkina2023evaluating,jin2023examining}, we measure the difference between training and test sets for chronological and random splits using two matrices: (i) Intersection over Union (IoU); (ii) DICE coefficient (DICE) \citep{dice1945measures}.

\textbf{Intersection over Union}

\begin{equation}
\footnotesize
    IoU = \frac{|V^{p} \cap V^{q}|}{|V^{p} \cup  V^{q}|}
\end{equation}

\textbf{DICE coefficient}

\begin{equation}
\footnotesize
    DICE = \frac{2 \times |V^{p} \cap V^{q}|}{|V^{p}| + |V^{q}|}
\end{equation}

where $V^{p}$ and $V^{q}$ denote the lists of unique tokens from two subsets (i.e., training and test sets) respectively. $|V^{p} \cap V^{q}|$ and $|V^{p} \cup  V^{q}|$ denote the total number of unique tokens that appear in the \textbf{intersection} and \textbf{union} of the two subsets respectively. When the two subsets have no shared vocabulary, the IoU and DICE values will be zero, while if they are identical, the IoU and DICE values will be equal to 1.

\subsection{Topics Drift}
\begin{figure}[!t]
\center
\includegraphics[width=0.48\textwidth]{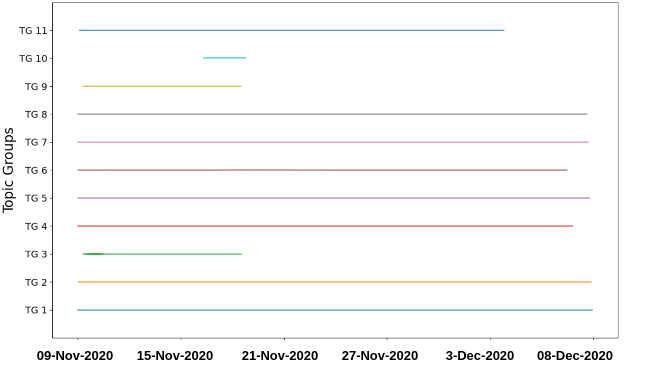}
\caption{Topic distribution over time from \citet{cotfas2021longest}. TG is short for `topic group'.}             
\label{fig:topic_distribution} 
\end{figure}

We also employ BERTopic \citep{grootendorst2022bertopic} to examine the temporal concept drift at the topic level on the Cotfas dataset \citep{cotfas2021longest}. We first extract the top 15 topic groups in the dataset using BERTopic. Then we manually combine similar topic groups and delete repeated topics and commonly used words (e.g., \textit{you}). Figure \ref{fig:topic_distribution} displays the distribution of 11 topic groups across the entire dataset over time. We observe that some topics only occur during certain periods, which indicates that using temporal data splits may yield an imbalanced topic distribution between the training and test sets. The phenomenon of topic drift may lead to a drop in model performance when using chronological data splits.

\subsection{Error Analysis}
We also manually conduct an analysis to investigate the model behaviours with two splitting strategies. First, when using random splits, we observe that some tweets in the test set are similar or identical to tweets in the training set using random splits rather than chronological splits (similar tweets are more likely to be generated during the same time period). This leads to a higher prediction accuracy when using random splits. For example, two pairs of  tweets from \citet{chen2022multilingual} and \citet{cotfas2021longest} are identical after data pre-processing:

\begin{quote}
    \footnotesize
    \textbf{\textcolor{black}{Tweet\_id:1220414** \& Tweet\_id:1220415**}} \textit{This deadly \#coronavirus was spread fr \#wuhan (China) to many countries just in days and kill 17 ppl. Maybe thousand ppl carrying pathogens are traveling around the worldIts spread fr human to human via air and still have no vaccinesPls read it; reduce the risk of infection. HTTPURL}
\end{quote}

\begin{quote} 
    \footnotesize
    \textbf{\textcolor{black}{Tweet\_id:133460** \& Tweet\_id:133520**:}} \textit{The U.S. \#airline industry and its pilots are essential to the distribution of a COVID-19 vaccine. Congress and government leaders must \#ExtendPSP now to ensure critical infrastructure is in place to distribute a vaccine—American lives depend on it @USER}
\end{quote}

Also, we observe that some tweets containing emerging topics (topics that appear in the later time period only) are correctly classified using random splits (topics overlap in both training and test sets) but wrongly using chronological splits. An example from \citet{mu2023vaxxhesitancy} is shown below.

\begin{quote}
\footnotesize
\textbf{\textcolor{black}{Tweet\_id:148891**}}: \textit{@USER \textbf{Vaccine passes} also impede freedom of movement. The irony. \emo{}{} \emo{}{} \emo{}{}}
\end{quote}
The data from \citet{mu2023vaxxhesitancy} covers from Nov 2020 to April 2022 while `Vaccine passes' was introduced in May 2021\footnote{\url{https://www.instituteforgovernment.org.uk/article/explainer/covid-passports}}. It is likely to cause models to fail to identify it in the testing set as models are unable to learn from the training set using chronological splits.

\begin{table}[!t]
\centering
\resizebox{\columnwidth}{!}{
\begin{tabular}{|ll|l|l|l|}
\hline
\multicolumn{1}{|l|}{\textbf{Datasets}} & \textbf{Splits} & \textbf{IoU} & \textbf{DICE} & \textbf{Acc} \\ \hline
\multicolumn{1}{|l|}{\multirow{2}{*}{\textbf{Cotfas}}} & \textit{Random}        & 0.17 & 0.25 & 85.1                   \\ \cline{2-5} 
\multicolumn{1}{|l|}{}                                 & \textit{Chronological} & 0.14 & 0.22 & 79.3                   \\ \hline
\multicolumn{1}{|l|}{\multirow{2}{*}{\textbf{Poddar}}}  & \textit{Random}        & 0.16 & 0.18 & 71.0                   \\ \cline{2-5} 
\multicolumn{1}{|l|}{}                                 & \textit{Chronological} & 0.14 & 0.17 & 70.5                   \\ \hline
\multicolumn{1}{|l|}{\multirow{2}{*}{\textbf{Mu}}}     & \textit{Random}        & 0.22 & 0.27 & 67.6                   \\ \cline{2-5} 
\multicolumn{1}{|l|}{}                                 & \textit{Chronological} & 0.22 & 0.26 & 68.3                   \\ \hline
\multicolumn{1}{|l|}{\multirow{2}{*}{\textbf{Di}}} & \textit{Random}        & 0.12 & 0.14 & 55.1                   \\ \cline{2-5} 
\multicolumn{1}{|l|}{}                                 & \textit{Chronological} & 0.11 & 0.13 & 55.0                   \\ \hline
\multicolumn{1}{|l|}{\multirow{2}{*}{\textbf{Chen}}}   & \textit{Random}        & 0.18 & 0.21 & 77.6                   \\ \cline{2-5} 
\multicolumn{1}{|l|}{}                                 & \textit{Chronological} & 0.17 & 0.20 & 75.5                   \\ \hline
\multicolumn{2}{|l|}{\textbf{Pearson coefficient}}                                       & \textbf{0.35} & \textbf{0.64} & \multicolumn{1}{c|}{-} \\ \hline
\end{tabular}
}
\caption{\small IoU and DICE values between training and test sets. \textbf{Acc} represents the best model accuracy. We also display the Pearson correlation between the two values and best accuracy values across all models.}
\label{tab:diff}
\end{table}

\section{Discussion}

\textbf{Q1 \& Q2: Chronological vs Random Splits in Multiple Languages}
In general, we notice that using random splits leads to an overestimation of performance compared to using chronological splits across the majority PLMs. Our findings align with the prior studies on temporal concept drift \citep{chalkidis2022improved,mu2023s}. However, previous work has shown the stance detection results are vulnerable to simple perturbations \citep{schiller2021stance}, which explains the results are not consistent over datasets from \citet{mu2023vaxxhesitancy} and \citet{di2022vaccineu} (the performance increases using chronological splits). Furthermore, we observe similar model performance for both data splitting strategies on \citet{mu2023vaxxhesitancy} and \citet{di2022vaccineu} datasets. Note that the results of the two distance measures (i.e., IoU and DICE) between the training and test set are also similar (Table \ref{tab:diff}).

\textbf{Q3: Vanilla vs. Domain Adapted PLMs}
We refer to the decrease in F1-score using chronological splits versus random splits as \textit{performance drop}. For all mono-lingual datasets, we observe that the performance drops less using domain-adapted PLMs (i.e., COVID-BERT and VAXX-BERT) than using vanilla BERT models. Taking COVID-BERT model for example, F1 scores decrease 10.8\% (-15.9\% for BERT), 1.7\% (-2.4\% for BERT) and even increase 0.5\% (-1.9\% for BERT) for datasets of \citet{cotfas2021longest}, \citet{poddar2022winds} and \citet{mu2023vaxxhesitancy} respectively (see Table \ref{tab:results_monolingual}). This indicates that applying domain \& task adaptation techniques can address the issue of temporal concept drift to a certain extent in stance detection towards COVID-19 related datasets. We also notice that f1 scores drop less using COVID-BERT than VAXX-BERT (e.g., -10.8\% vs. -14.7\% on the data set of \citet{cotfas2021longest}). We speculate that this is because a more domain-specific model (i.e., VAXX-BERT) lead to poorer generalise ability and is less sensitive to time.


\textbf{Q4: Distance Between Training and Test Sets}
In table \ref{tab:diff}, we observed that using random splits results in significantly higher IoU and DICE scores (note that higher scores indicate greater similarities between the training and test sets) compared to chronological splits.
This suggests that new topics (i.e., temporal concept drift) emerge in the test sets when using the chronological split strategy.
Also, we discover a positive Pearson correlation between the model accuracy and the similarity distance of two subsets using both IoU (0.35) and DICE (0.64) metrics, i.e., the higher the values, the higher the model accuracy.

\section{Conclusion}
We explored how temporalities affect stance detection towards COVID-19 vaccination on Twitter. Our experiments showed that using chronological splits significantly reduces the accuracy of stance classification in existing datasets. Therefore, we believe that developing real-world stance detection approaches should take temporal factors into account. Meanwhile, our results suggest that using domain- and task-adaptive models, and combining models trained on different time periods, can effectively address the effects of temporal concept drift in COVID-19 vaccination stance detection.

\section*{Ethics Statement}
Our work has been approved by the Research Ethics Committee of our institute, and complies with the
policies of Twitter API. 
All datasets are publicly available via the links (see footnotes) provided in the original papers.
\section*{Acknowledgements}
This research is supported by a UKRI grant EP/W011212/1 (``XAIvsDisinfo: eXplainable AI Methods for Categorisation and Analysis of COVID-19 Vaccine Disinformation and Online Debates'')\footnote{\url{https://gow.epsrc.ukri.org/NGBOViewGrant.aspx?GrantRef=EP/W011212/1}} and an EU Horizon 2020 grant (agreement no.871042) (``So-BigData++: European Integrated Infrastructure for Social Mining and BigData Analytics'')\footnote{\url{http://www.sobigdata.eu}}.

\section*{References}
\bibliographystyle{lrec-coling2024-natbib}
\bibliography{lrec-coling2024-example}

\begin{thebibliography}{27}
\expandafter\ifx\csname natexlab\endcsname\relax\def\natexlab#1{#1}\fi

\bibitem[{Alkhalifa et~al.(2021)Alkhalifa, Kochkina, and Zubiaga}]{alkhalifa2021opinions}
Rabab Alkhalifa, Elena Kochkina, and Arkaitz Zubiaga. 2021.
\newblock Opinions are made to be changed: Temporally adaptive stance classification.
\newblock In \emph{Proceedings of the 2021 workshop on open challenges in online social networks}, pages 27--32.

\bibitem[{Alkhalifa et~al.(2023)Alkhalifa, Kochkina, and Zubiaga}]{alkhalifa2023building}
Rabab Alkhalifa, Elena Kochkina, and Arkaitz Zubiaga. 2023.
\newblock Building for tomorrow: Assessing the temporal persistence of text classifiers.
\newblock \emph{Information Processing \& Management}, 60(2):103200.

\bibitem[{Chalkidis and S{\o}gaard(2022)}]{chalkidis2022improved}
Ilias Chalkidis and Anders S{\o}gaard. 2022.
\newblock Improved multi-label classification under temporal concept drift: Rethinking group-robust algorithms in a label-wise setting.
\newblock In \emph{Findings of the Association for Computational Linguistics: ACL 2022}, pages 2441--2454.

\bibitem[{Chen et~al.(2022)Chen, Chen, and Pang}]{chen2022multilingual}
Ninghan Chen, Xihui Chen, and Jun Pang. 2022.
\newblock A multilingual dataset of covid-19 vaccination attitudes on twitter.
\newblock \emph{Data in Brief}, 44:108503.

\bibitem[{Ciotti et~al.(2020)Ciotti, Ciccozzi, Terrinoni, Jiang, Wang, and Bernardini}]{ciotti2020covid}
Marco Ciotti, Massimo Ciccozzi, Alessandro Terrinoni, Wen-Can Jiang, Cheng-Bin Wang, and Sergio Bernardini. 2020.
\newblock The covid-19 pandemic.
\newblock \emph{Critical reviews in clinical laboratory Sci.}, 57(6):365--388.

\bibitem[{Conneau et~al.(2020)Conneau, Khandelwal, Goyal, Chaudhary, Wenzek, Guzm{\'a}n, Grave, Ott, Zettlemoyer, and Stoyanov}]{conneau2020unsupervised}
Alexis Conneau, Kartikay Khandelwal, Naman Goyal, Vishrav Chaudhary, Guillaume Wenzek, Francisco Guzm{\'a}n, {\'E}douard Grave, Myle Ott, Luke Zettlemoyer, and Veselin Stoyanov. 2020.
\newblock Unsupervised cross-lingual representation learning at scale.
\newblock In \emph{Proceedings of the 58th Annual Meeting of the Association for Computational Linguistics}, pages 8440--8451.

\bibitem[{Cotfas et~al.(2021)Cotfas, Delcea, Roxin, Ioan{\u{a}}{\c{s}}, Gherai, and Tajariol}]{cotfas2021longest}
Liviu-Adrian Cotfas, Camelia Delcea, Ioan Roxin, Corina Ioan{\u{a}}{\c{s}}, Dana~Simona Gherai, and Federico Tajariol. 2021.
\newblock The longest month: analyzing covid-19 vaccination opinions dynamics from tweets in the month following the first vaccine announcement.
\newblock \emph{Ieee Access}, 9:33203--33223.

\bibitem[{Devlin et~al.(2019)Devlin, Chang, Lee, and Toutanova}]{devlin2019bert}
Jacob Devlin, Ming-Wei Chang, Kenton Lee, and Kristina Toutanova. 2019.
\newblock Bert: Pre-training of deep bidirectional transformers for language understanding.
\newblock In \emph{Proceedings of the 2019 Conference of the North American Chapter of the Association for Computational Linguistics: Human Language Technologies, Volume 1 (Long and Short Papers)}, pages 4171--4186.

\bibitem[{Di~Giovanni et~al.(2022)Di~Giovanni, Pierri, Torres-Lugo, and Brambilla}]{di2022vaccineu}
Marco Di~Giovanni, Francesco Pierri, Christopher Torres-Lugo, and Marco Brambilla. 2022.
\newblock Vaccineu: Covid-19 vaccine conversations on twitter in french, german and italian.
\newblock In \emph{Proceedings of the International AAAI Conference on Web and Social Media}, volume~16, pages 1236--1244.

\bibitem[{Dice(1945)}]{dice1945measures}
Lee~R Dice. 1945.
\newblock {Measures of the Amount of Ecologic Association Between Species}.
\newblock \emph{Ecology}, 26(3):297--302.

\bibitem[{Florio et~al.(2020)Florio, Basile, Polignano, Basile, and Patti}]{florio2020time}
Komal Florio, Valerio Basile, Marco Polignano, Pierpaolo Basile, and Viviana Patti. 2020.
\newblock {Time of your hate: The challenge of time in hate speech detection on social media}.
\newblock \emph{Applied Sciences}, 10(12):4180.

\bibitem[{Glandt et~al.(2021)Glandt, Khanal, Li, Caragea, and Caragea}]{glandt2021stance}
Kyle Glandt, Sarthak Khanal, Yingjie Li, Doina Caragea, and Cornelia Caragea. 2021.
\newblock Stance detection in covid-19 tweets.
\newblock In \emph{Proceedings of the 59th Annual Meeting of the Association for Computational Linguistics and the 11th International Joint Conference on Natural Language Processing (Volume 1: Long Papers)}, pages 1596--1611.

\bibitem[{Grootendorst(2022)}]{grootendorst2022bertopic}
Maarten Grootendorst. 2022.
\newblock Bertopic: Neural topic modeling with a class-based tf-idf procedure.
\newblock \emph{arXiv preprint arXiv:2203.05794}.

\bibitem[{Hu et~al.(2023)Hu, Sheng, Cao, Zhu, Wang, Wang, and Jin}]{hu-etal-2023-learn}
Beizhe Hu, Qiang Sheng, Juan Cao, Yongchun Zhu, Danding Wang, Zhengjia Wang, and Zhiwei Jin. 2023.
\newblock Learn over past, evolve for future: Forecasting temporal trends for fake news detection.
\newblock In \emph{Proceedings of the 61st Annual Meeting of the Association for Computational Linguistics (Volume 5: Industry Track)}, pages 116--125, Toronto, Canada. Association for Computational Linguistics.

\bibitem[{Huang and Paul(2019)}]{huang2019neural}
Xiaolei Huang and Michael Paul. 2019.
\newblock Neural temporality adaptation for document classification: Diachronic word embeddings and domain adaptation models.
\newblock In \emph{Proceedings of the 57th Annual Meeting of the Association for Computational Linguistics}, pages 4113--4123.

\bibitem[{Jin et~al.(2023)Jin, Mu, Maynard, and Bontcheva}]{jin2023examining}
Mali Jin, Yida Mu, Diana Maynard, and Kalina Bontcheva. 2023.
\newblock Examining temporal bias in abusive language detection.
\newblock \emph{arXiv preprint arXiv:2309.14146}.

\bibitem[{Kochkina et~al.(2023)Kochkina, Hossain, Logan~IV, Arana-Catania, Procter, Zubiaga, Singh, He, and Liakata}]{kochkina2023evaluating}
Elena Kochkina, Tamanna Hossain, Robert~L Logan~IV, Miguel Arana-Catania, Rob Procter, Arkaitz Zubiaga, Sameer Singh, Yulan He, and Maria Liakata. 2023.
\newblock Evaluating the generalisability of neural rumour verification models.
\newblock \emph{Information Processing \& Management}, 60(1):103116.

\bibitem[{Lopez~Bernal et~al.(2021)Lopez~Bernal, Andrews, Gower, Gallagher, Simmons, Thelwall, Stowe, Tessier, Groves, Dabrera et~al.}]{lopez2021effectiveness}
Jamie Lopez~Bernal, Nick Andrews, Charlotte Gower, Eileen Gallagher, Ruth Simmons, Simon Thelwall, Julia Stowe, Elise Tessier, Natalie Groves, Gavin Dabrera, et~al. 2021.
\newblock Effectiveness of covid-19 vaccines against the b. 1.617. 2 (delta) variant.
\newblock \emph{New England Journal of Medicine}, 385(7):585--594.

\bibitem[{Mu et~al.(2023{\natexlab{a}})Mu, Bontcheva, and Aletras}]{mu2023s}
Yida Mu, Kalina Bontcheva, and Nikolaos Aletras. 2023{\natexlab{a}}.
\newblock It’s about time: Rethinking evaluation on rumor detection benchmarks using chronological splits.
\newblock In \emph{Findings of the Association for Computational Linguistics: EACL 2023}, pages 724--731.

\bibitem[{Mu et~al.(2023{\natexlab{b}})Mu, Jin, Grimshaw, Scarton, Bontcheva, and Song}]{mu2023vaxxhesitancy}
Yida Mu, Mali Jin, Charlie Grimshaw, Carolina Scarton, Kalina Bontcheva, and Xingyi Song. 2023{\natexlab{b}}.
\newblock Vaxxhesitancy: A dataset for studying hesitancy towards covid-19 vaccination on twitter.
\newblock In \emph{Proceedings of the International AAAI Conference on Web and Social Media}, volume~17, pages 1052--1062.

\bibitem[{Mu et~al.(2023{\natexlab{c}})Mu, Song, Bontcheva, and Aletras}]{mu2023examining2}
Yida Mu, Xingyi Song, Kalina Bontcheva, and Nikolaos Aletras. 2023{\natexlab{c}}.
\newblock Examining the limitations of computational rumor detection models trained on static datasets.
\newblock \emph{arXiv preprint arXiv:2309.11576}.

\bibitem[{M{\"u}ller et~al.(2023)M{\"u}ller, Salath{\'e}, and Kummervold}]{muller2020covid}
Martin M{\"u}ller, Marcel Salath{\'e}, and Per~E Kummervold. 2023.
\newblock Covid-twitter-bert: A natural language processing model to analyse covid-19 content on twitter.
\newblock \emph{Frontiers in Artificial Intelligence}, 6:1023281.

\bibitem[{Poddar et~al.(2022{\natexlab{a}})Poddar, Mondal, Misra, Ganguly, and Ghosh}]{poddar2022winds}
Soham Poddar, Mainack Mondal, Janardan Misra, Niloy Ganguly, and Saptarshi Ghosh. 2022{\natexlab{a}}.
\newblock Winds of change: Impact of covid-19 on vaccine-related opinions of twitter users.
\newblock In \emph{Proceedings of the International AAAI Conference on Web and Social Media}, volume~16, pages 782--793.

\bibitem[{Poddar et~al.(2022{\natexlab{b}})Poddar, Samad, Mukherjee, Ganguly, and Ghosh}]{CAVES}
Soham Poddar, Azlaan~Mustafa Samad, Rajdeep Mukherjee, Niloy Ganguly, and Saptarshi Ghosh. 2022{\natexlab{b}}.
\newblock Caves: A dataset to facilitate explainable classification and summarization of concerns towards covid vaccines.
\newblock In \emph{Proceedings of the 45th International ACM SIGIR Conference on Research and Development in Information Retrieval}, pages 3154--3164.

\bibitem[{Schiller et~al.(2021)Schiller, Daxenberger, and Gurevych}]{schiller2021stance}
Benjamin Schiller, Johannes Daxenberger, and Iryna Gurevych. 2021.
\newblock {Stance Detection Benchmark: How Robust is Your Stance Detection?}
\newblock \emph{KI-K{\"u}nstliche Intelligenz}, pages 1--13.

\bibitem[{Wilkinson et~al.(2016)Wilkinson, Dumontier et~al.}]{wilkinson2016fair}
Mark~D Wilkinson, Michel Dumontier, et~al. 2016.
\newblock The fair guiding principles for scientific data management and stewardship.
\newblock \emph{Scientific data}, 3(1):1--9.

\bibitem[{Xing et~al.(2018)Xing, Cambria, and Welsch}]{xing2018natural}
Frank~Z Xing, Erik Cambria, and Roy~E Welsch. 2018.
\newblock Natural language based financial forecasting: a survey.
\newblock \emph{Artificial Intelligence Review}, 50(1):49--73.

\end{thebibliography}


\end{document}